\documentclass[journal,10pt]{IEEEtran}

\usepackage{graphicx}
\usepackage{amsmath}
\usepackage{subfigure}
\usepackage{multirow}
\usepackage{cite}

\begin{document}

\title{Research on Mosaic Image Data Enhancement for Overlapping Ship Targets}

\author{Guangmiao Zeng, Wanneng Yu, Rongjie Wang$ ^{*} $ and Anhui Lin

\thanks{R. Wang is with the School of Marine Engineering, Jimei University, Xiamen, 3610211, China e-mail:roger811207@163.com}
\thanks{Manuscript received XXX, XX, 2015; revised XXX, XX, 2015.}}


\maketitle

\begin{abstract}
The problem of overlapping occlusion in target recognition has been a difficult research problem, and the situation of mutual occlusion of ship targets in narrow waters still exists. In this paper, an improved mosaic data enhancement method is proposed, which optimizes the reading method of the data set, strengthens the learning ability of the detection algorithm for local features, improves the recognition accuracy of overlapping targets while keeping the test speed unchanged, reduces the decay rate of recognition ability under different resolutions, and strengthens the robustness of the algorithm. The real test experiments prove that, relative to the original algorithm, the improved algorithm improves the recognition accuracy of overlapping targets by 2.5\%, reduces the target loss time by 17\%, and improves the recognition stability under different video resolutions by 27.01\%.
\end{abstract}


\IEEEpeerreviewmaketitle

\section{Introduction}

At present, with the rapid development of computer technology, image recognition technology does not only stay in the laboratory, but has been widely used in various fields in the society. And the urgent problem of sea surface target recognition is one of them, as an important part in the development of ship intelligent navigation technology. In recent years, there are many recognition methods for sea surface targets, most of which are remote sensing images of sea surface obtained by synthetic aperture radar. Guo et al. improved the network structure of CenterNet to enhance the recognition of small targets of ships \cite{1}, Li et al. proposed a new two-branch regression network to improve the localization of ships \cite{2}, and Fu et al. used an anchor-free frame approach with feature balancing and refinement networks to improve the detection of ships in complex scenes \cite{3}. The above-mentioned articles all identify sea surface ship targets in vertical view and need to be photographed by aircraft, satellites and other aerial vehicles, while there are fewer studies related to identification of surrounding sea surface targets by horizontal view on board ships. Not only that, most of the studies have discussed the problem of small target identification, which takes into account the relatively long field of view in wide waters, but there are also situations where ships are sailing in narrow waters, when there are more ships around, and it is very easy for them to block each other in the observation field of view, so the problem of quickly and accurately locating targets with different degrees of overlap for identification becomes one of the key points.

There have been some research results on the problem of recognition of obscured targets. Wan et al. recovered local facial features through generative adversarial networks to achieve face recognition under occlusion \cite{4}, Chowdhury et al. combined progressive expansion algorithms with graph attention networks to improve license plate recognition under street congestion \cite{5}, and Liu et al. used coupled networks to improve recognition accuracy for small targets or occluded pedestrians \cite{6}. They improved the recognition of obscured targets by optimizing the neural network, which more or less increases the computational complexity and reduces the detection speed of target recognition due to the enhancements made to the network structure.

At this stage, target detection methods are mainly divided into two types, the first one is a two-stage detection method represented by the R-CNN series \cite{7,8,9,10}, which first extracts the region of interest using the region proposal network, and then traverses the test images from top to bottom and from left to right for recognition and detection based on the size of the region of interest. The second one is the one-stage detection algorithm represented by SSD \cite{11} and YOLO series \cite{12,13,14,15}, which does not rely on the region proposal network, but directly uses the anchor frame, multiple square regions into which the test image is segmented, and recognizes and detects each region separately. In contrast, the one-stage method is more accurate but generates a certain amount of redundant computations. The two-stage method greatly improves the detection speed at the expense of a portion of accuracy, which is beneficial to meet the needs of real-time monitoring.

The Yolov4 algorithm \cite{15} will be used as the basic model for the experiments, and although it has been proposed for a relatively short period of time, it has already been used in agriculture \cite{16}, construction \cite{17}, medicine \cite{18} and other fields. The experiments in the aforementioned literature were conducted using servers deployed on land for training and testing; the situation is different in that the communication signal at sea is relatively unstable and the size and energy consumption of the recognition system needs to be minimized due to the complex sea conditions, so the detection system needs to be considered for offline use. Therefore, the Yolov4-tiny lightweight algorithm is chosen as the main experimental model in this paper.

Based on the above analysis, an improved mosaic data enhancement method is proposed in this paper, and the main contributions can be summarized as follows.

(1) Data enhancement is performed on the images in the training dataset to enhance the recognition capability of the target recognition algorithm for overlapping targets of ships without changing the network structure.

(2) For the small mobile platform in offline state, a suitable lightweight algorithm is selected to improve the recognition accuracy while maintaining its recognition speed, and to reduce the impact of input images with different resolutions on the recognition performance of the algorithm.

(3) By conducting real-time detection tests in different sea areas of the ship dataset, the results show that the algorithm trained by the improved mosaic method is more accurate in the detection of overlapping targets of ships, which proves the effectiveness and robustness of the improved method.

The remainder of this paper is organized as follows: Section II describes in detail the main structures of the improved moasic method and the Yolov4-tiny algorithm. Section III describes the simulation experiment comparison and the results of the real test experiment. Finally, conclusions are drawn in Section IV.

\section{Methodology}

The Yolo \cite{12} algorithm has received a lot of attention since it was first proposed in 2016, and the subsequently proposed Yolov2 \cite{13} and Yolov3 \cite{14} algorithms have made a series of improvements to the model structure, data preprocessing methods, and loss function calculation methods based on it, which greatly improve the speed and accuracy of target detection. And the Yolov4 algorithm adds many optimization techniques to the calculation method of Yolov3 algorithm, which improves the recognition accuracy with the same recognition speed.

\subsection{Overview of the Yolov4-tiny algorithm}

The Yolov4-tiny network is based on the Yolov4 network simplified by reducing the number of parameters by a factor of 10 at the expense of some recognition accuracy. It is reduced from about 60 million parameters in the Yolov4 network to about 6 million parameters in Yolov4-tiny. Its network structure is shown in Figure 1.

\begin{figure}
	\centering\includegraphics[width=3.3in]{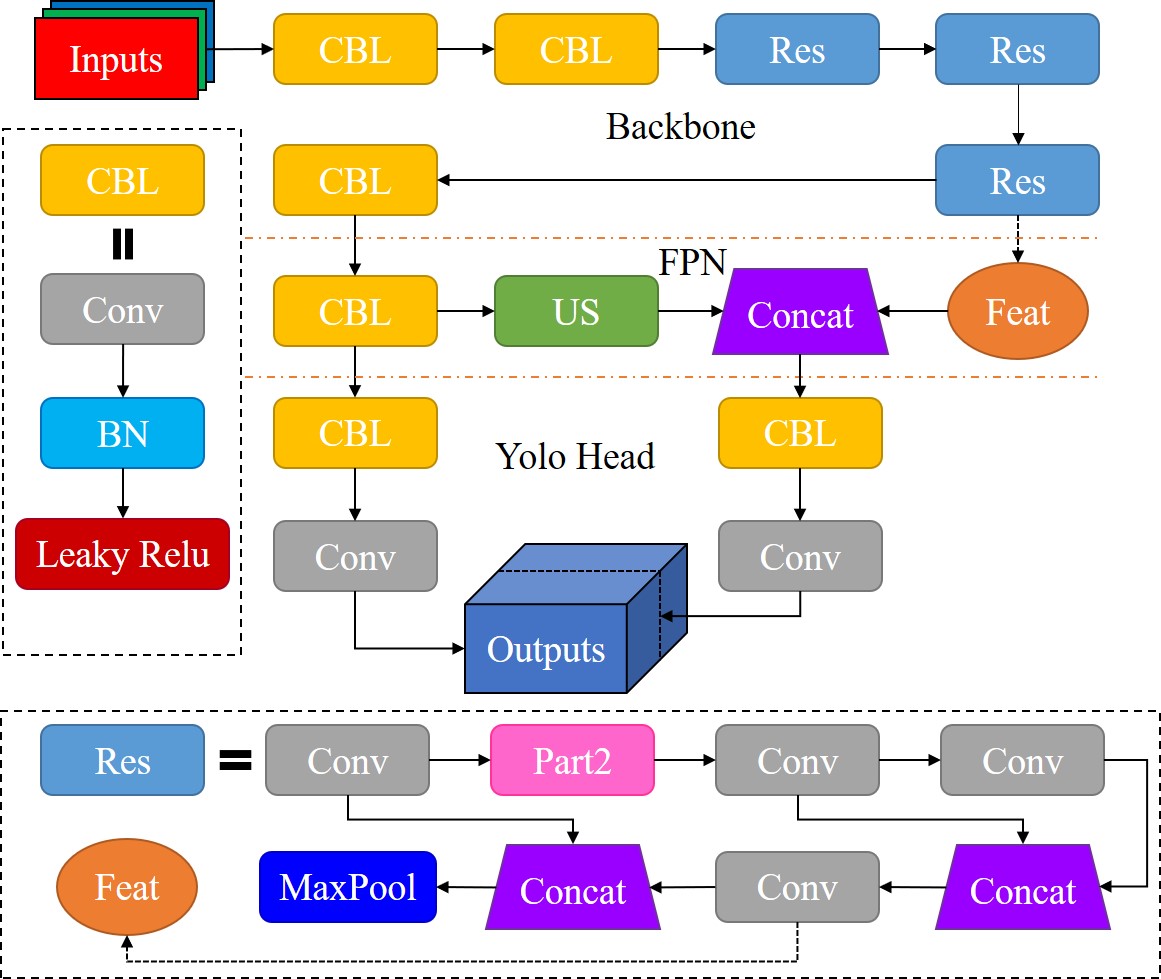}
	\caption{Characteristic structure of Yolov4-tiny network}
\end{figure}

Among them, the convolutional block in the backbone network consists of convolutional layers, batch normalization layers \cite{19}, and leaky Relu \cite{20} activation function. And the residual block is the CSPDarknet53-Tiny networks \cite{21}, and its structure is shown in Figure 2.

\begin{figure}
	\centering\includegraphics[width=3.3in]{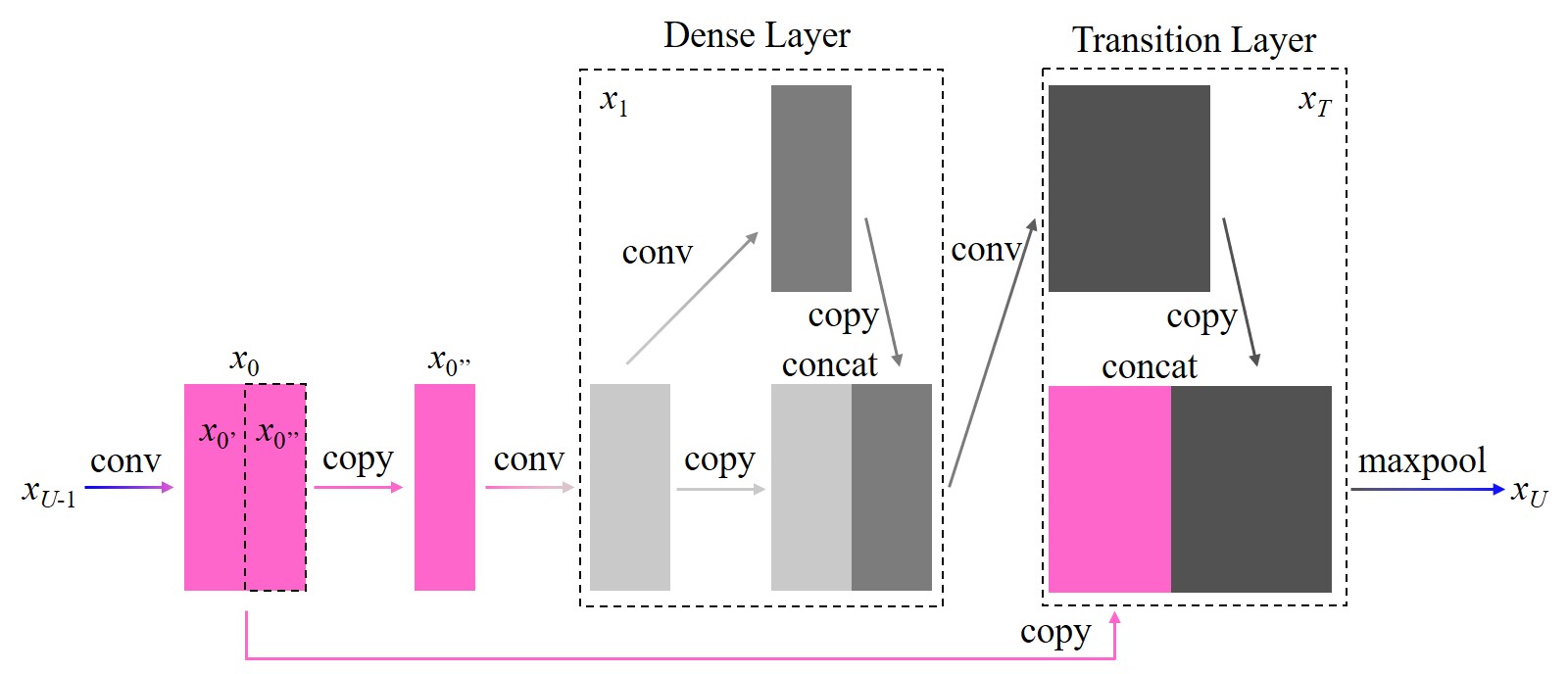}
	\caption{Structure of residual blocks in CSPDarknet53-Tiny networks}
\end{figure}

It consists of a dense layer and a connected layer, which first performs a convolution operation on the output $ x_{U-1} $ of the previous convolution layer to generate a new convolution layer, and divides its output $ x_0 $ = $ [x_{0^{'}},x_{0^{''}}] $  into two parts $ x_{0^{'}} $ and $ x_{0^{''}} $ before and after for the forward propagation. In the network structure of Yolov4-tiny, the second part is taken first for forward propagation, and then the first part is directly concatenated with the second part to the end of the stage, skipping the dense layer. Waiting until the $ x_{0^{''}} $ of the second part finishes the forward calculation after feature concatenation with $ x_0 $ in the transition layer, the output $ x_T $ is obtained, which undergoes max pooling to produce the output $ x_U $ of the residual block. The process of forward propagation and backward propagation of the residual block is shown in Eqs. (1) to (2).

\begin{align} \begin{split}
	x_T &= w_T \cdot \left[ x_{0^{''}} , x_1 \right] \\
	x_U &= w_U \cdot \left[ x_0 , x_1 \right]
\end{split} \end{align}

\begin{align} \begin{split}
	\omega ^{'}_{T} &= f_T \left( \omega _T \cdot \left\{ g_{0}^{''} , g_1 \right\} \right) \\
	\omega ^{'}_{U} &= f_U \left( \omega _U \cdot \left\{ g_0 , g_T \right\} \right)
\end{split} \end{align}

\noindent where $\omega_{i}$ and $\omega^{'}_{i}$ are the weights during forward and backward propagation, $ f_{i} $ denotes the function of weight update, and $ g_{i} $ denotes the gradient propagated to the $ i^{th} $ layer. $ i $ equals T or U, representing the output of the connected layer or residual block, respectively. Therefore, using the structure of CSPNet in back propagation, the gradients on different channels can be integrated separately, for example, when the gradient information passes through the dense layer, it will only change the weights on the $ x_{0^{''}} $ channel but will not affect $ x_{0^{'}} $. This reduces the excessive and repetitive gradient information while retaining the feature values at different depths, reduces the memory overhead and improves the network computation speed without affecting the network feature extraction effect.

After the backbone network, the network features are optimized using the feature pyramid structure \cite{22} , and the implementation of a small feature pyramid for the Yolov4-tiny network is shown in Figure 3.

\begin{figure}
	\centering\includegraphics[width=2.3in]{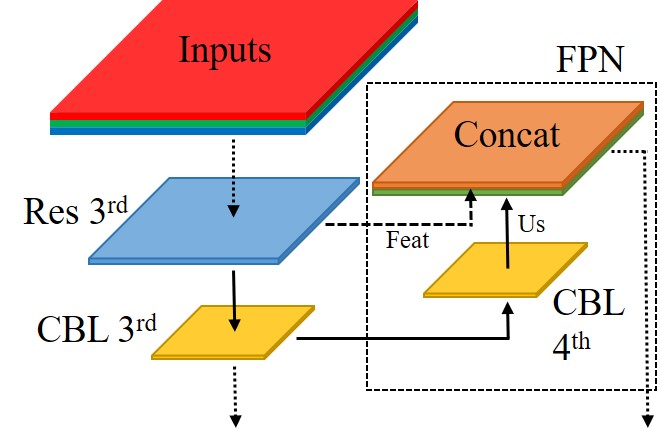}
	\caption{Structure of a small feature pyramid}
\end{figure}

The intermediate features are the output of the fourth convolutional layer in the third residual block of the backbone network, which is concatenated with the up-sampled features from the output of the fourth convolutional block in the network. As can be seen in Figure 1, the backbone network contains only the first three CBL layers, so the output generated by the concatenation operation performed in Figure 3 represents the fusion of the shallow network features and the deep network features. Due to the top-down feature extraction by multi-layer convolution, the deep network retains most of the feature values of large targets, and few or even zero of the feature values of small targets are preserved. Therefore, the feature pyramid structure is used to extract the features of several different layers of the network, and after up-sampling and amplification, they are stitched together from the bottom up to achieve the feature fusion of multiple layers, which improves the recognition ability of the network for different size targets at multiple separate rates.

After that, the two outputs of the small feature pyramid are plugged into the head network for calculation, and two sets of images containing different perceptual fields are generated, which are adjusted to the prior frame contained in themselves respectively. The non-maximum suppression method is used to identify and detect targets of different sizes in the original image and improve the overall detection capability of the neural network for multi-scale targets.

\subsection{Data enhancement methods}

The mosaic method is an extension of the CutMix \cite{23} method to generate a new data enhancement algorithm, which differs from the two-image overlay fusion of the CutMix method; instead, it uses four images for cropping and stitching to form a new image. This method can better enrich the background of the target and prevent the degradation of the network generalization ability due to the similar background of the training set.

The output image of Yolov4-tiny algorithm contains two different perceptual fields, while the output image of Yolov4 algorithm has three different perceptual fields, so the recognition ability of Yolov4-tiny algorithm for multi-scale targets will be relatively weak. Therefore, it is especially important to improve the data enhancement method to enhance the generalization ability of the network.

The improved mosaic data enhancement method is shown in Figure 4. The original mosaic method uses the top and middle orange channels in Figure 2 for feature enhancement, while the improved mosaic method adds the bottom gold channel to the original one and uses three channels for feature enhancement. The output of the third golden channel is obtained by increasing the number of images arranged in each row and column compared to the above two. For the sake of illustration, the newly generated nine-in-one image with a specification of $ 3*3 $ will be called $ m9 $, the four-in-one image with a specification of $ 2*2 $ will be called $ m4 $, and the image generated without merging with a specification of $ 1*1 $ will be called $ m1 $. The ratio of $ m1 $, $ m4 $ and $ m9 $ is $ o:p:q $. This combination, to some extent, makes the scale variation characteristics of the training dataset more diverse, thus further attenuating the interference of the background on the target features.

\begin{figure}
	\centering\includegraphics[width=2.3in]{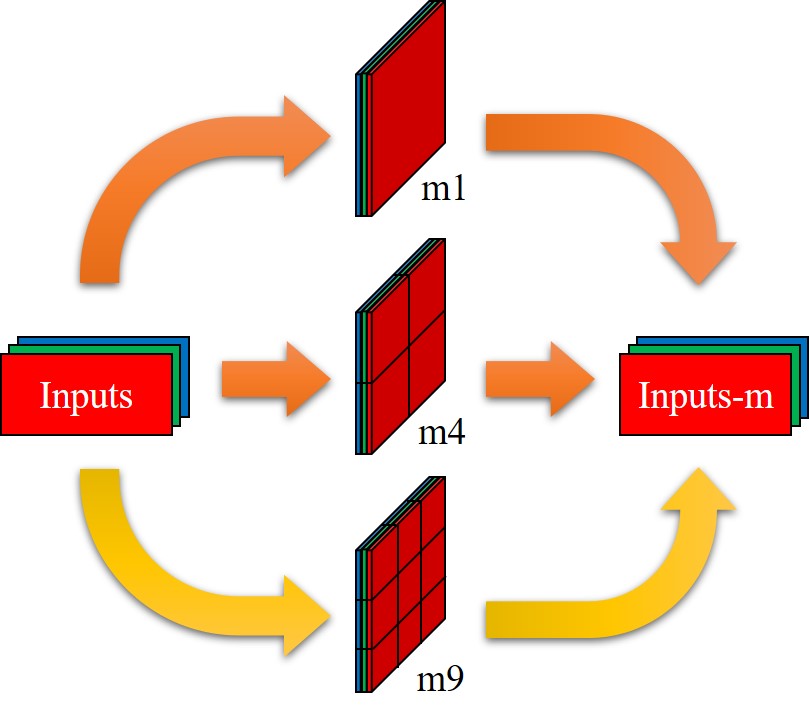}
	\caption{The improved mosaic data enhancement method}
\end{figure}

The nine-in-one image $ m9 $ is generated as shown in Fig. 5, which is mainly divided into three stages \textbf{A}, \textbf{B}, and \textbf{C}. In stage \textbf{A}, the width and height ($ W $, $ H $) of the input image are used as the boundary values, and the image is first scaled with the scaling multipliers $ t_{X} $ and $ t_{Y} $ for the X and Y axes, as shown in Eqs. (3)-(4).

\begin{align}
	{t_X} &= {f_{rand}}({t_W},{t_W} + \Delta {t_W})\\
	{t_{\mathop{\rm Y}\nolimits} } &= {f_{rand}}({t_H},{t_H} + \Delta {t_H})
\end{align}

\noindent where $ t_{W} $ and $ t_{H} $ are the minimum values of the wide and high scaling multipliers, respectively, and $ \Delta t_{W} $ and  $ \Delta t_{H} $ are the lengths of the random intervals of the wide and high scaling multipliers, respectively, both of which are hyperparameters. $ f_{rand}() $ denotes the random value function.

\begin{figure}
	\centering\includegraphics[width=3.3in]{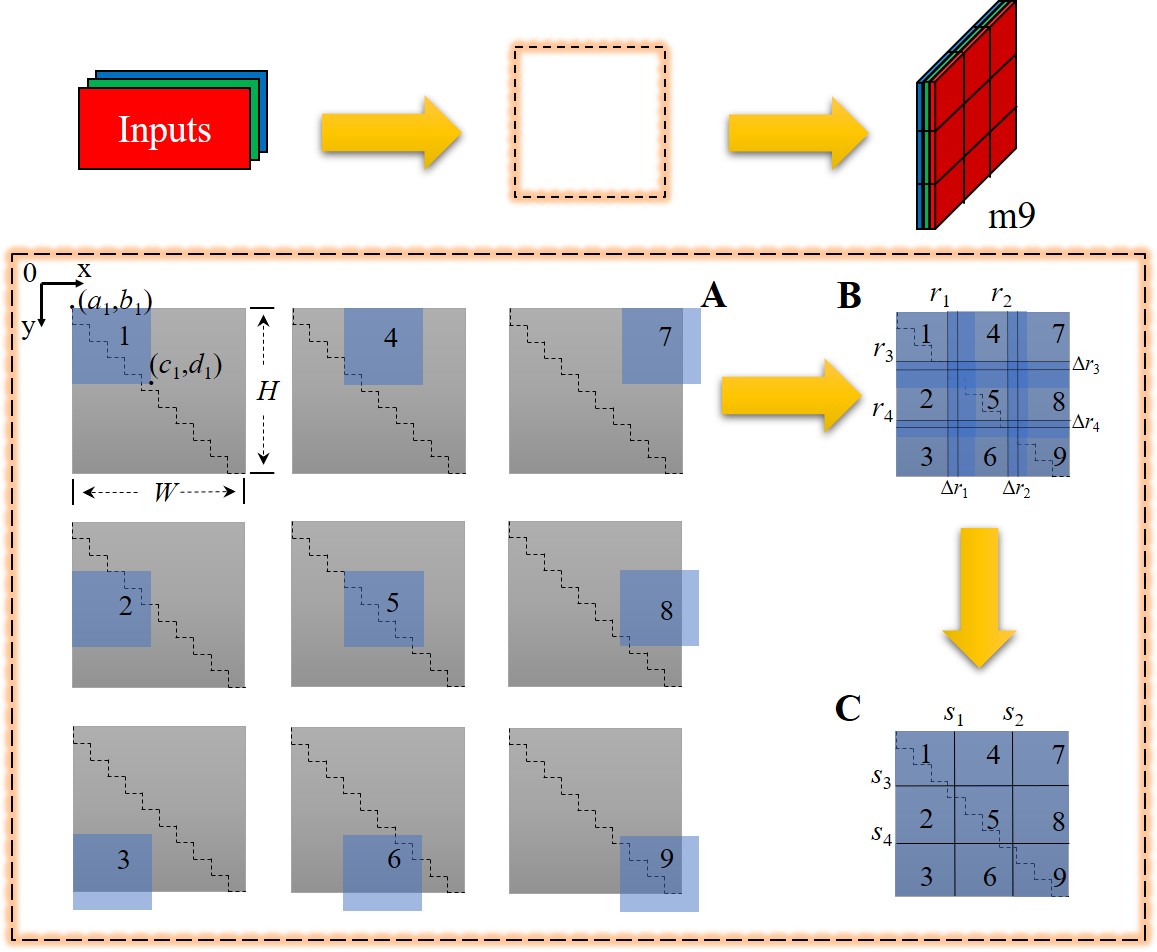}
	\caption{Flowchart of $ m9 $ image generation}
\end{figure}

The coordinates of the top-left and bottom-right corners of the image after scaling are $ [(a_{i}, b_{i}), (c_{i}, d_{i})] $, which are obtained from Eqs. (5)-(8).

\begin{align}
	a_i = \begin{cases}
		0, & i = 1,2,3 \\
		W \cdot r_1, & i = 4,5,6 \\
		W \cdot r_2, & i = 7,8,9
		  \end{cases} \\
	b_i = \begin{cases}
		0, & i = 1,4,7 \\
		H \cdot r_3, & i = 2,5,8 \\
		H \cdot r_4, & i = 3,6,9
		  \end{cases}
\end{align}

\begin{align}
	{c_i} &= {a_i} + W \cdot {t_W}\\
	{d_i} &= {b_i} + H \cdot {t_H}
\end{align}

\noindent Among them, $ r_{1} $ and $ r_{2} $ are the ratio of the distance between the upper left coordinate point and the 0 point of the two sets of images except the 0 point of the X-axis to the total width, respectively, and $ r_{3} $ and $ r_{4} $ are the ratio of the distance between the upper left coordinate point and the 0 point of the two sets of images except the 0 point of the Y-axis to the total height, respectively, and both are also hyperparameters. And the short black lines in the gray area are the scale bars, each small segment represents a one-tenth of the width or height. Using the scale bars, we can see that the scaling of the images from the 2nd to the 9th sheet is the same as that of the 1st sheet, and the width and height are both $ t_{W} $  and $ t_{H} $ times of the original.

In stage \textbf{B}, the nine images cropped in the previous stage need to be stitched together and the part of the overflowing bounding box cropped off. It can be seen that there is a certain degree of overlap in the merged images, so each small area needs to be divided. From the schematic diagram of stage \textbf{A}, it can be seen that when the scaled images are placed at the specified position according to the coordinates, there will be an overflowing border. At this time, the overflowing part needs to be cropped, as shown in Eqs. (9)-(10).

\begin{align}
	c_{i}^{'} = \begin{cases}
		c_i, & \text{if } c_1 < W \\
		W, & \text{if } c_1 \ge W
				\end{cases} \\
	c_{i}^{'} = \begin{cases}
		d_i, & \text{if } d_1 < H \\
		H, & \text{if } d_1 \ge H
				\end{cases} 
\end{align}

After edge cropping, the four square regions enclosed by eight two-by-two parallel dotted lines are used as random intervals of the split lines. Where the value of $ r_{i}=(r_{1}, r_{2}, r_{3}, r_{4}) $ is equal to the ratio of the distance between the coordinates of the split line and the point 0 to the length of the boundary, and $ \Delta i $ is the length of the random interval of the split line.

In stage \textbf{C}, a second cut will be made to the internal overlapping part, whose split line coordinates $ s_{i} $ can be obtained from Eq. (11)

\begin{equation}
	{s_i} = {f_{rand}}({r_i},{r_i} + \Delta {r_i})\;\;\;i = 1,2,3,4
\end{equation}

After cropping, the stitched $ m9 $ image is obtained. Since the original image is partially missing in the scaling and stitching process, it is possible that the targets at the edges of the original image are partially or completely cut off during the operation. Therefore, it is also necessary to crop or even reject the real frame corresponding to these targets to meet the needs of target detection.

\begin{figure}
	\centering\includegraphics[width=3.3in]{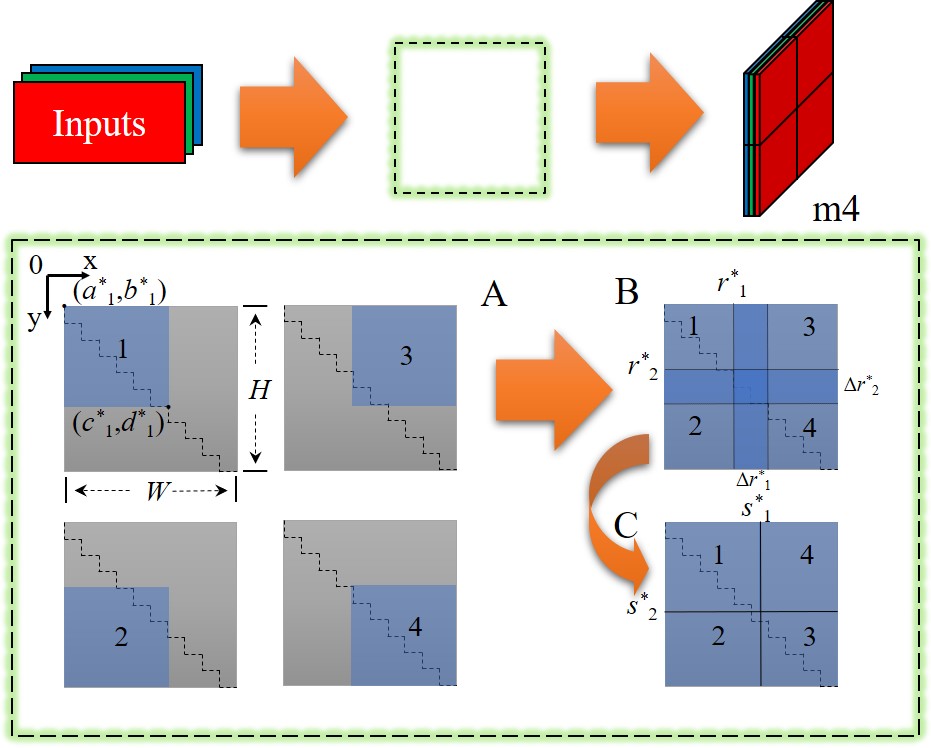}
	\caption{Flowchart of $ m4 $ image generation}
\end{figure}

The method of generating the $ m4 $ image is similar to that of generating $ m9 $, as shown in Figure 6. The upper left corner coordinates $ (a^{*}_{i}, b^{*}_{i}) $ and the split line coordinates $ s^{*}_{i} $ are shown by Eqs. (12)-(14).

\begin{align}
	a^{*}_{i} = \begin{cases}
		0 , & i = 1,2 \\
		W \cdot r^{*}_{1} , & i = 3,4
				\end{cases}  \\
	b^{*}_{i} = \begin{cases}
		0 , & i = 1,3 \\
		H \cdot r^{*}_{2} , & i = 2,4
				\end{cases}
\end{align}

\begin{equation}
	s^{*}_{i} = f_{rand} \left( r^{*}_{i} , r^{*}_{i} + \Delta r^{*}_{i} \right) , i = 1,2
\end{equation}

\noindent where $ ^{*} $ is used as the distinguishing symbol between the $ m4 $ image and the $ m9 $ image. The scaling multipliers $ t^{*}_{X} $, $ t^{*}_{Y} $, $ \Delta t^{*}_{W} $ and $ \Delta t^{*}_{H} $ and the lower right corner coordinates $ (c^{*}_{i}, d^{*}_{i}) $ are calculated in the same way as for the $ m9 $ image. Since there is no out-of-border overflow, only the internal overlap part needs to be segmented and cropped.

The $ m1 $ image only requires feature enhancement by conventional methods such as flip and color gamut change before entering the network because there is no stitching of multiple images.

\subsection{Loss function}
The Yolov4-tiny algorithm as a whole is roughly the same as its modified previous version, and its loss function contains three components: loss of confidence in the target ($ Loss_{\rm{conf}} $), loss of classification ($ Loss_{\rm{cls}} $), and loss of position ($ Loss_{\rm{loc}} $). As shown in Eqs. (15)-(19).

\begin{gather}
	\begin{split}
	Loss = \lambda_{\rm{conf}} \cdot Loss_{\rm{conf}} + \lambda _{\rm{cls}} \cdot Loss_{\rm{cls}} \\
	+ \lambda _{\rm{loc}} \cdot Loss_{\rm{loc}} 
	\end{split} \\
	\begin{split}
	Loss_{\rm{conf}} = \sum\limits_{i = 0}^{K \times K} \sum\limits_{j = 0}^M I_{ij}^{\rm{obj}} Loss_{\rm{BCE}} \left( \hat C_i , C_i \right)  \\ 
	- \sum\limits_{i = 0}^{K \times K} \sum\limits_{j = 0}^M I_{ij}^{\rm{noobj}} Loss_{\rm{BCE}} \left( \hat C_i , C_i \right) 
	\end{split} \\
	Loss_{\rm{cls}} = \sum\limits_{i = 0}^{K \times K} I_{ij}^{\rm{obj}} \sum\limits_{k \in \rm{classes}}^{K \times K} Loss_{\rm{BCE}} \left( \hat p_i(k) , p_i(k) \right)  \\
	Loss_{\rm{BCE}} ( \hat N , N) = \hat N \log (N) + (1 - \hat N) \log (1 - N) \\
	Loss_{\rm{loc}} = \sum\limits_{i = 0}^{K \times K} \sum\limits_{j = 0}^M I_{ij}^{\rm{obj}} \cdot loss_{\rm{CIoU}}
\end{gather}

\noindent where $ \lambda_{conf} $, $ \lambda_{cls} $ and $ \lambda_{loc} $ represent the weights of three different categories of loss in the loss function, respectively, the Yolov4-tiny network divides each input image into K × K cells first, and each grid produces M anchor boxes. After each anchor is subjected to the network's antecedent computation, an adjusted bounding box is obtained, and the total number of anchors is K × K × M. $ I^{obj}_{ij} $ and $ I^{noobj}_{ij} $ are used to determine whether the center coordinates of the target are in the $ j^{th} $ anchor box in the $ i^{th} $ grid, if yes the former is equal to 1 and the latter is equal to 0, otherwise the opposite. $ C_{i} $ is the confidence of the true box in the $ i^{th} $ cell and  $ {\hat C_i} $ is the confidence of the prediction box in the $ i^{th} $ cell. $ p_{i}(k) $ denotes the conditional probability that the true box in the $ i^{th} $ cell contains the $ k^{th} $ type of target and $ {\hat p_i}(k) $ denotes the conditional probability that the prediction box in the $ i^{th} $ cell contains the $ k^{th} $ type of target.

Unlike the Yolov3 algorithm, the Yolov4-tiny algorithm uses CIoU loss in the calculation of the location loss function \cite{24} instead of the cross-entropy loss used in the confidence loss and classification loss, which enables a more accurate description of the location information. The CIoU loss is calculated as shown in Eqs. (20)-(24).

\begin{gather}
	loss_{\rm{CIoU}} = 1 - IoU + R_{\rm{CIoU}} \left(  B , B ^{gt} \right) \\
	IoU = \frac{ \left| B \cap B^{gt} \right| }{ \left| B \cup B^{gt} \right| } \\
	R_{\rm{CIoU}} \left( \rm B ,\rm B^{gt} \right) = \frac{\rho \left( \rm b , \rm b^{gt} \right)}{c^2} + \alpha v \\
	\alpha  = \frac{v}{\left( 1 - IoU \right) + v} \\
	v = \frac{4}{\pi ^2} \left( \arctan \frac{w^{gt}}{h^{gt}} - \arctan \frac{w}{h} \right)^2
\end{gather}

\noindent where $ IoU $ is the intersection ratio, the prediction box $ B = (x, y, w, h) $, and the true box $ B^{gt} = (x^{gt}, y^{gt}, w^{gt}, h^{gt}) $, which consist of x, y coordinates indicating the location of the center point and w, h coordinates indicating the width and height length. $ R_{CIoU}(B, B^{gt}) $ is the penalty term between the prediction box $ B $ and the real box $ B^{gt} $, $ b $ and $ b^{gt} $ represent the centroids of $ B $ and $ B^{gt} $, $ \rho(\bullet) $ denotes the Euclidean distance, and $ c $ is the diagonal distance of the smallest box that can contain both the prediction box and the real box. $ \alpha $ is a positive trade-off parameter and $ v $ is a parameter that measures the consistency of the aspect ratio, which gives a higher priority to factors in the region where the predicted box overlaps with the true box relative to the non-overlapping part in the regression calculation.

\subsection{Network optimization methods}
In order to better combine the characteristics of the ship dataset, the Yolov4-tiny algorithm uses the K-mean clustering algorithm to divide the real frames of different sizes in the training set into m classes before starting the training, and the boxes represented by the center points of the real boxes in each class are used as anchor boxes, so that anchor boxes can be obtained that are more suitable for detecting ship targets. In this paper m=6, these anchor boxes will be divided into 2 groups of 3 boxes each according to the size to detect target objects of different scales.

In the convolution block, the data extracted by the convolution layer, after batch normalization, is activated using the Leaky Relu activation function, which does not set all negative values to 0 as in the Relu function, but sets a non-zero slope, as shown in Eq. (25).

\begin{align}
	f_{\rm{Leaky Relu}}(n_i) = \begin{cases}
		n_i, & \text{if } n_i \ge 0 \\
		\varphi \cdot n_i, & \text{if } n_i < 0, a_i \in (0,1)
	\end{cases}
\end{align}

\noindent where $ \varphi $ is the slope when the input value is less than 0 and is the hyperparameter.

In the early stage of training, using a large learning rate can make the network converge quickly, while in the later stage of training, using a small learning rate is more also beneficial for the network to converge to the optimal value. Therefore, the exponential decay strategy of learning rate is utilized for training, and the learning rate $ \gamma $ is calculated as shown in equation (26).

\begin{equation}
	\gamma  = \varepsilon ^\tau \gamma_0
\end{equation}

\noindent where $ \gamma_0 $ denotes the initial learning rate, $ \varepsilon $ is the decay rate, and $ \tau $ is the number of iterations of the training network.

The Yolov4-tiny model is built based on convolutional neural networks, so its features extracted at different depth levels are not the same. So the network model is first trained in a large dataset, and when it has the ability to extract basic and abstract features, it is then fine-tuned using transfer learning methods to transfer the weights and biases after training to the network in the new training environment. Since the types of targets are different in various training sets, the weights and deviations of the last layer of the network model are structured differently, except that they can be transferred.

\section{Experimental simulation and testing}

\subsection{Planning of data sets}

The widely used datasets, such as VOC dataset \cite{25} and COCO dataset \cite{26}, contain ships that are classified into only 1 category, which exist with random image size and low resolution (no more than 640*360). In contrast, the ship dataset \cite{27} used for training and testing in this paper contains a total of 7000 images with a resolution of 1920*1080, which are intercepted from video clips taken by surveillance cameras that belong to a sea surface surveillance system deployed along the coastline, which includes 156 cameras in 50 different locations. The dataset contains six different types of ships, the number and categories of which are shown in Table 1.

\begin{table}[]
	\caption{Ship data set target object category and number}
	\centering
	\begin{tabular}{cc}
		\hline
		\textbf{Vessel type} & \textbf{Quantity(pcs)} \\ \hline
		Ore carrier          & 2084                   \\
		Bulk cargo carrier   & 1811                   \\
		General cargo ship   & 1426                   \\
		Container ship       & 898                    \\
		Fishing boat         & 1539                   \\
		Passenger ship       & 455                    \\ \hline
	\end{tabular}
\end{table}

The ships in these images have different lighting conditions, observation angles, distance and proximity scales, and overlap levels, making the dataset much more complex and increasing the difficulty of target detection algorithm recognition.

\subsection{Training and test results}

The algorithms in this paper were implemented on the open source neural network framework Pytorch (3.8.5). The computational workstation configuration consists of a GPU (GeForce RTX 3090), CPU (AMD Ryzen 9 3950x 16 Core/ 3.5 GHz/72 M), and 128 G RAM. It consists of a 1080p camera module, power supply module, display output module, and control module, as shown in Figure 7.

\begin{figure}
	\centering\includegraphics[width=3.0in]{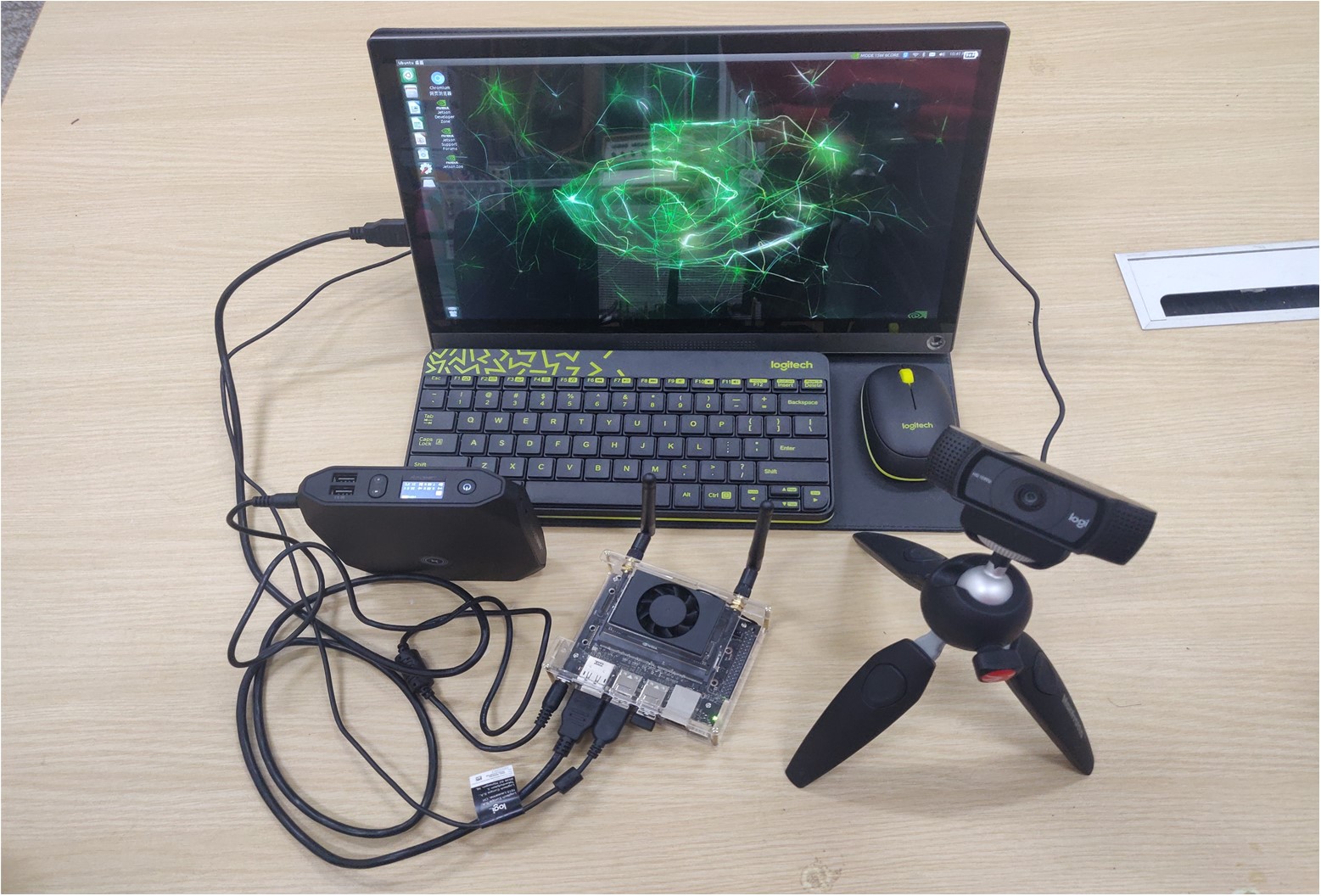}
	\caption{Small mobile test platform}
\end{figure}

Before the training starts, the dataset is classified. 6000 target images of ships with almost no occlusion are selected as the training set, and 1000 images of ships with different severity of overlap occur as the test set. By optimizing the target recognition algorithm, the target ships with different degrees of overlap and occlusion can be captured more quickly and accurately, reducing and improving the recognition accuracy.The parameters of the mosaic method in the experiment are shown in Table 2, and the parameters in the network optimization method are shown in Table 3.

\begin{table}[]
	\caption{Experimental parameters in mosaic data enhancement methods}
	\centering
	\resizebox{\columnwidth}{!}{
	\begin{tabular}{cccccc}
		\hline
		\textbf{Parameter} & \textbf{Numeric} & \textbf{Parameter} & \textbf{Numeric} & \textbf{Parameter} & \textbf{Numeric} \\ \hline
		$ W $              & 608              & $ \Delta r3 $      & 0.05             & $ \Delta r^{*}_{1} $               & 0.2              \\
		$ H $              & 608              & $ \Delta r4 $      & 0.05             & $ \Delta r^{*}_{2} $               & 0.2              \\
		$ r_{1}  $         & 0.3              & $ t_{W} $          & 0.4              & $ t^{*}_{W} $               & 0.4              \\
		$ r_{2} $          & 0.65             & $ t_{H}  $         & 0.4              & $ t^{*}_{H} $               & 0.4              \\
		$ r_{3} $          & 0.3              & $ \Delta t_{W} $   & 0.05             & $ \Delta t^{*}_{W}  $    & 0.2              \\
		$ r_{4}  $         & 0.65             & $ \Delta t_{H} $   & 0.05             & $ \Delta t^{*}_{H} $               & 0.2              \\
		$ \Delta r_{1} $   & 0.05             & $ r^{*}_{1} $      & 0.4              &                   &                  \\
		$ \Delta r_{2} $   & 0.05             & $ r^{*}_{2} $      & 0.4              &                    &                  \\ \hline
	\end{tabular}
	}
\end{table}

\begin{table}
	\caption{Parameters in network optimization methods}
	\centering
	\begin{tabular}{cc}
		\hline
		\textbf{Parameter} & \textbf{Numeric} \\ \hline
		$ \varphi $        & 0.1              \\
		$ \gamma _{0}  $   & $ 1 \times 10^{-3} $ \\
		$ \varepsilon $    & 0.95             \\ \hline
	\end{tabular}
\end{table}

The images used for training and validation in the training dataset are randomly segmented in a ratio of 9 to 1. After training starts, the network stops after 100 iterations, and Figure 8 represents the recognition accuracy of the Yolov4-tiny algorithm at different numbers of iterations. The values of the legend in the figure are the values of o:p:q, and the accuracy is represented by mAP.

As can be seen in Figure 8, the values of the yellow curve are slightly higher than the values of the green curve during the last twenty iterations that tend to be smooth, and the recognition results of the Yolov4-tiny algorithm improve slightly after using the mosaic method, while the improved mosaic method indicated by the blue curve greatly improves the recognition accuracy, even higher than that of the Yolov4 using the original mosaic method algorithm using the original mosaic method. So the improved mosaic method not only improves the recognition accuracy of the Yolov4-tiny algorithm, but also greatly improves the detection speed of ship recognition compared to the Yolov4 algorithm. The experimental procedure of data enhancement using m4 and m9 methods is shown in Figure 9.

\begin{figure*}
	\centering\includegraphics[width=0.7\linewidth]{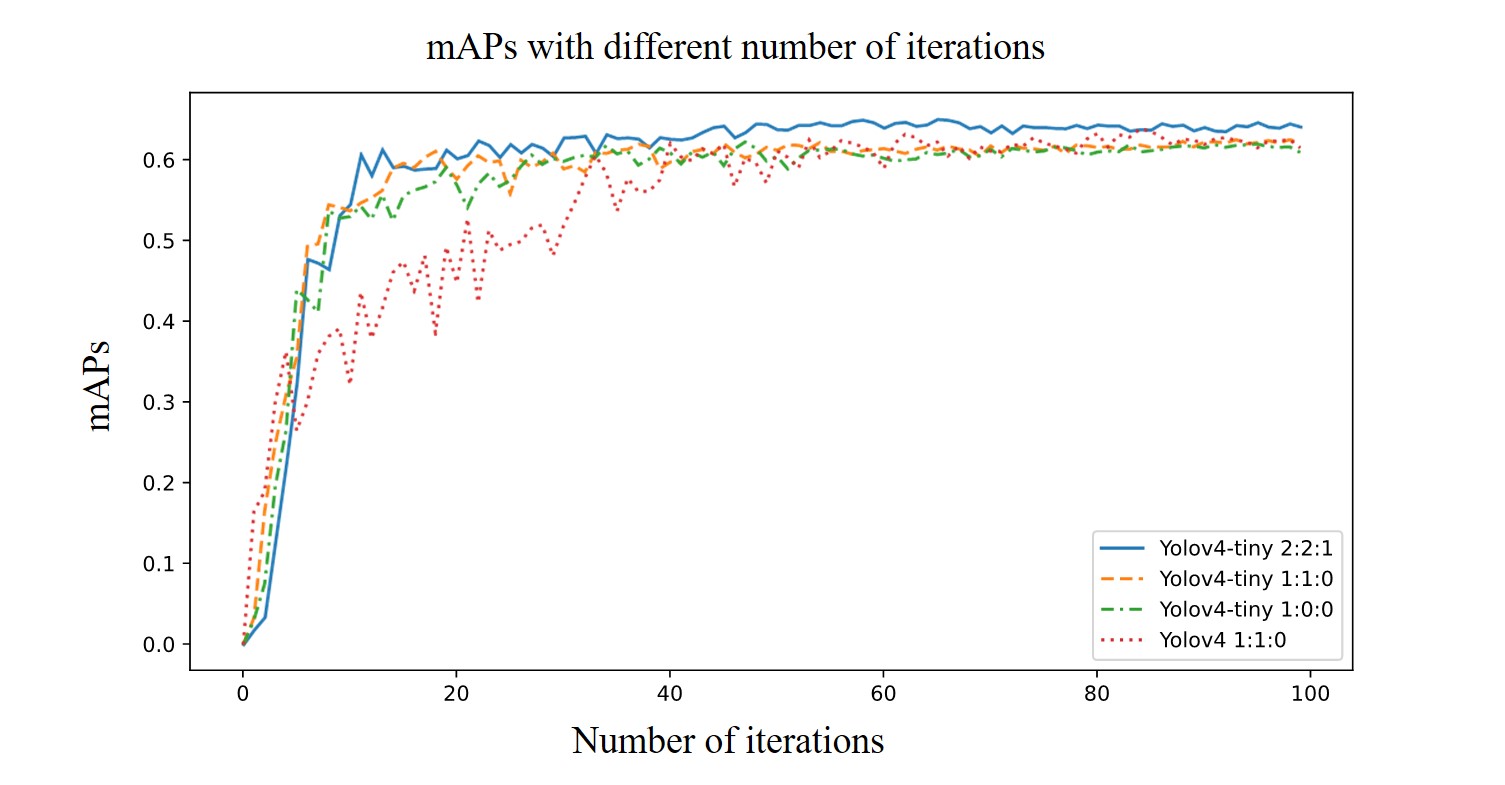}
	\caption{mAPs of Yolov4-tiny algorithm with different number of iterations}
\end{figure*}

\begin{figure}
	\centering
	\subfigure[]{
	\begin{minipage}[b]{0.4\linewidth}
		\includegraphics[width=1\textwidth]{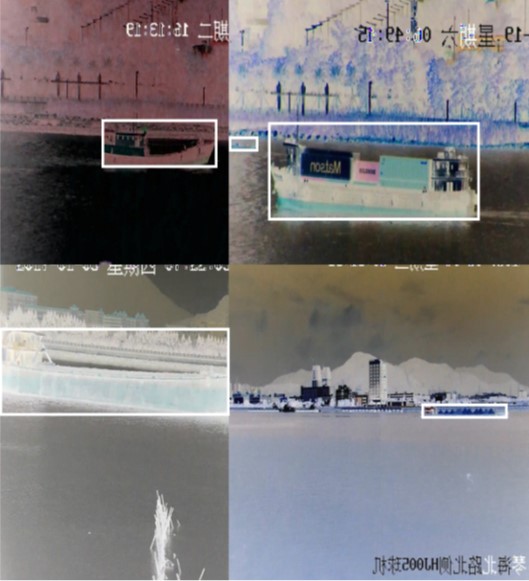}
	\end{minipage}
	}
	\subfigure[]{
	\begin{minipage}[b]{0.4\linewidth}
		\centering
		\includegraphics[width=1\textwidth]{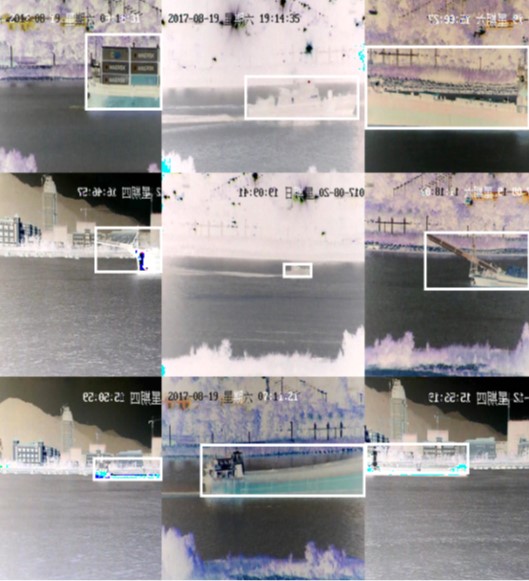}
	\end{minipage}
	}
	\centering
	\caption{Experimental procedure of $ m4 $ and $ m9 $ methods, (a) $ m4 $, (b) $ m9 $}
\end{figure}

The curve with the highest recognition accuracy in Figure 8 was trained using the modified mosaic method, which uses the $ m1 $, $ m4 $ and $ m9 $ methods as inputs with probability according to the ratio $ o:p:q $ = 2:2:1 . In order to investigate the effect of the different ratios on the recognition accuracy, several sets of experiments were conducted in this paper to compare the results as shown in Table 4.

\begin{table*}
	\caption{Effect of mosaic data enhancement methods on target recognition accuracy at different scales}
	\centering
	\begin{tabular}{ccccc}
		\hline
		\textbf{Algorithm Type}      & \textbf{o:p:q} & \textbf{Highest mAP in 100 iterations} & \textbf{Average mAP in the last 20   iterations} & \textbf{Ratio relative to the original   mosaic (1:1:0)} \\ \hline
		\multirow{9}{*}{Yolov4-tiny} & 1:0:0          & 62.28\%                                & 61.54\%                                          & 99.27\%                                                  \\
		& 1:1:0          & 62.56\%                                & 61.99\%                                          & 100.00\%                                                 \\
		& 1:1:1          & 62.39\%                                & 60.53\%                                          & 97.64\%                                                  \\
		& 1:2:1          & 61.88\%                                & 60.10\%                                          & 96.95\%                                                  \\
		& 2:1:1          & 63.11\%                                & 61.44\%                                          & 99.11\%                                                  \\
		& \textbf{2:2:1} & \textbf{65.06\%}                       & \textbf{64.09\%}                                 & \textbf{103.39\%}                                        \\
		& 3:2:1          & 63.78\%                                & 62.49\%                                          & 100.81\%                                                 \\
		& 4:2:1          & 63.31\%                                & 62.86\%                                          & 101.40\%                                                 \\
		& 4:3:2          & 62.98\%                                & 62.25\%                                          & 100.42\%                                                 \\ \hline
		Yolov4                       & 1:1:0          & 63.90\%                                & 62.56\%                                          & 100.92\%                                          \\ \hline                                                
	\end{tabular}
\end{table*}

In Table 4, it can be seen that relative to the original Mosaic (1:1:0) method, the recognition algorithm will achieve better results when $ m4 $ is satisfied with double $ m9 $ and $ m1 $ is greater than or equal to $ m4 $, and its recognition accuracy will be higher than the Yolov4 algorithm applying the original Mosaic method when $ o:p:q $ = 2:2:1 and 4:2:1. In addition, not all the improved methods at all ratios are superior, which indicates that the dataset should be focused on $ m1 $ data, so that the network learns the overall features of the target well, and on this basis, $ m4 $ and $ m9 $ are used to enhance the learning of local features, respectively, to improve the generalization ability of the network.

To further test the recognition capability of the network for overlapping targets, a small mobile testbed is used for real-time inspection of the sea surface, located at Gulangyu Island (Xiamen, China). A real-time video clip of two fishing boats overlapping during travel was used as the experiment to calculate the recognition capability of the network at each moment. The video clip had a resolution of 1080P, a duration of 38 seconds, 24 frames per second, and a total of 912 frames, and the test results are shown in Figure 10. The weight file selected in the actual test experiment is the weight corresponding to the highest mAP in 100 iterations.

\begin{figure*}
	\centering\includegraphics[width=0.8\linewidth]{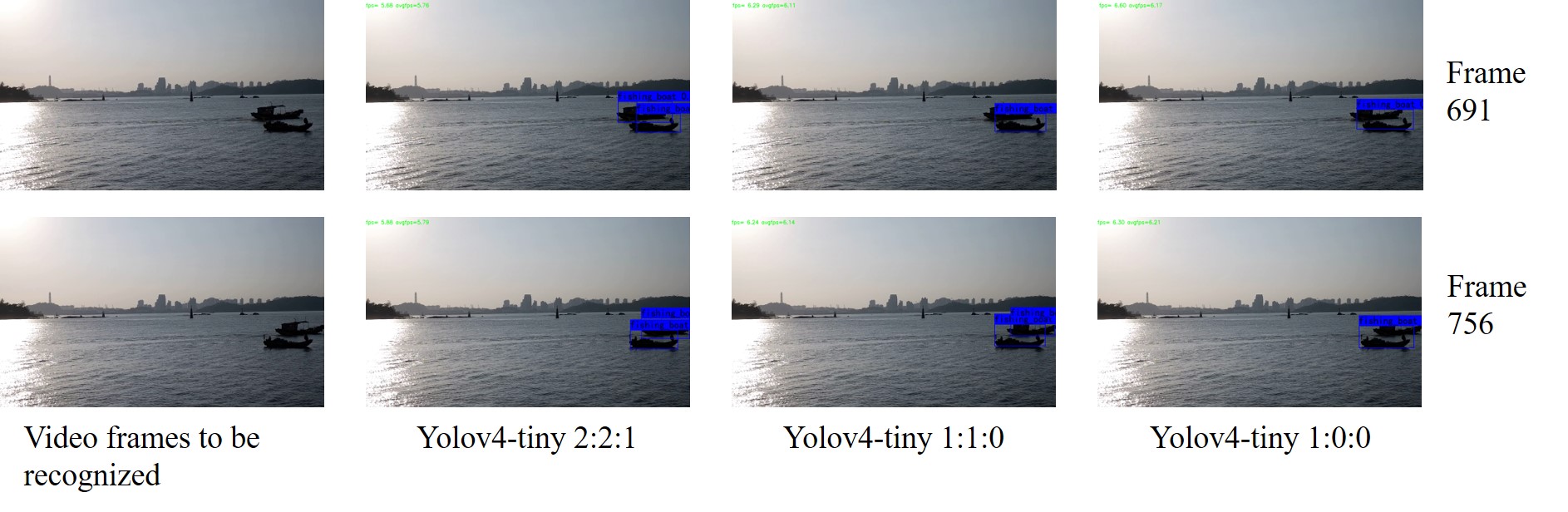}
	\caption{Overlap detection experiment comparison chart}
\end{figure*}

In Figure 10, it can be seen that in frame 691, when the two boats are already severely overlapped, the Yolov4-tiny algorithm using the improved mosaic method can still identify both boats at the same time, while the Yolov4-tiny algorithm using the original mosaic method and the one not using the mosaic method can only identify one of the boats. In frame 756 when the two ships are getting rid of the overlap, the Yolov4-tiny algorithm with the improved mosaic method and the original mosaic method can already recognize both ships at the same time, while the Yolov4-tiny algorithm without the mosaic method can still recognize only one of the ships. This further illustrates the effectiveness of the improved mosaic method in improving the detection capability of overlapping targets.

Due to the limited arithmetic power of small mobile devices, the recognition speed is slow at 1080p resolution, so the detection of ship targets in different resolution videos is tested and the recognition speed of the network is shown in Table 5.

\begin{table*}[]
	\caption{Comparison of recognition speed at different resolutions}
	\centering
	\begin{tabular}{ccccccc}
		\hline
		\multirow{2}{*}{\textbf{Network Type}} & \multicolumn{4}{c}{\textbf{Recognition speed /fps}}            & \multirow{2}{*}{\textbf{Number of network weights/pcs}} & \multirow{2}{*}{\textbf{Network weight size /MB}} \\
		& \textbf{1080p} & \textbf{720p} & \textbf{480p} & \textbf{360p} &                                                         &                                                   \\ \hline
		Yolov4-tiny                            & 6.34           & 8.68          & 10.48         & 11.38         & 5,885,666                                               & 22.45                                             \\
		Yolov4                                 & 2.03           & 2.27          & 2.41          & 2.46          & 63,964,611                                              & 244.01  \\ \hline                                         
	\end{tabular}
\end{table*}

\begin{table*}
	\caption{Comparison of the recognition ability of three mosaic methods for overlapping problems at different resolutions}
	\centering
	\resizebox{\linewidth}{!}{
		\begin{tabular}{lllllll}
			\hline
			\textbf{Video Resolution} & \textbf{o:p:q} & \textbf{\begin{tabular}[c]{@{}l@{}}Unable to separate \\ the starting frames\end{tabular}} & \textbf{\begin{tabular}[c]{@{}l@{}}Unable to separate \\ end frames\end{tabular}} & \textbf{\begin{tabular}[c]{@{}l@{}}Number of frames that \\ cannot be separated/pcs\end{tabular}} & \textbf{\begin{tabular}[c]{@{}l@{}}Unable to separate \\ time/sec\end{tabular}} & \textbf{\begin{tabular}[c]{@{}l@{}}Proportion of time notseparable\\ relative to the original \\ Mosaic method (1:1:0)\end{tabular}} \\ \hline
			\multirow{3}{*}{1080p}    & 1:0:0          & 654                                                                                        & 804                                                                               & 151                                                                                               & 6.29                                                                            & 196\%                                                                                                                                \\
			& 1:1:0          & 676                                                                                        & 752                                                                               & 77                                                                                                & 3.21                                                                            & 100\%                                                                                                                                \\
			& \textbf{2:2:1} & \textbf{692}                                                                               & \textbf{755}                                                                      & \textbf{64}                                                                                       & \textbf{2.67}                                                                   & \textbf{83\%}                                                                                                                        \\ \hline
			\multirow{3}{*}{720p}     & 1:0:0          & 649                                                                                        & 810                                                                               & 162                                                                                               & 6.75                                                                            & 210\%                                                                                                                                \\
			& 1:1:0          & 673                                                                                        & 752                                                                               & 80                                                                                                & 3.33                                                                            & 104\%                                                                                                                                \\
			& \textbf{2:2:1} & \textbf{691}                                                                               & \textbf{755}                                                                      & \textbf{65}                                                                                       & \textbf{2.71}                                                                   & \textbf{84\%}                                                                                                                        \\ \hline
			\multirow{3}{*}{480p}     & 1:0:0          & 650                                                                                        & 912                                                                               & 263                                                                                               & 10.96                                                                           & 342\%                                                                                                                                \\
			& 1:1:0          & 658                                                                                        & 753                                                                               & 96                                                                                                & 4.00                                                                            & 125\%                                                                                                                                \\
			& \textbf{2:2:1} & \textbf{692}                                                                               & \textbf{755}                                                                      & \textbf{64}                                                                                       & \textbf{2.67}                                                                   & \textbf{83\%}                                                                                                                        \\ \hline
			\multirow{3}{*}{360p}     & 1:0:0          & 649                                                                                        & 912                                                                               & 264                                                                                               & 11.00                                                                           & 343\%                                                                                                                                \\
			& 1:1:0          & 656                                                                                        & 754                                                                               & 99                                                                                                & 4.13                                                                            & 129\%                                                                                                                                \\
			& \textbf{2:2:1} & \textbf{692}                                                                               & \textbf{756}                                                                      & \textbf{65}                                                                                       & \textbf{2.71}                                                                   & \textbf{84\%}                                                                                                                        \\ \hline
		\end{tabular}
	}
\end{table*}

\begin{figure*}
	\centering\includegraphics[width=0.7\linewidth]{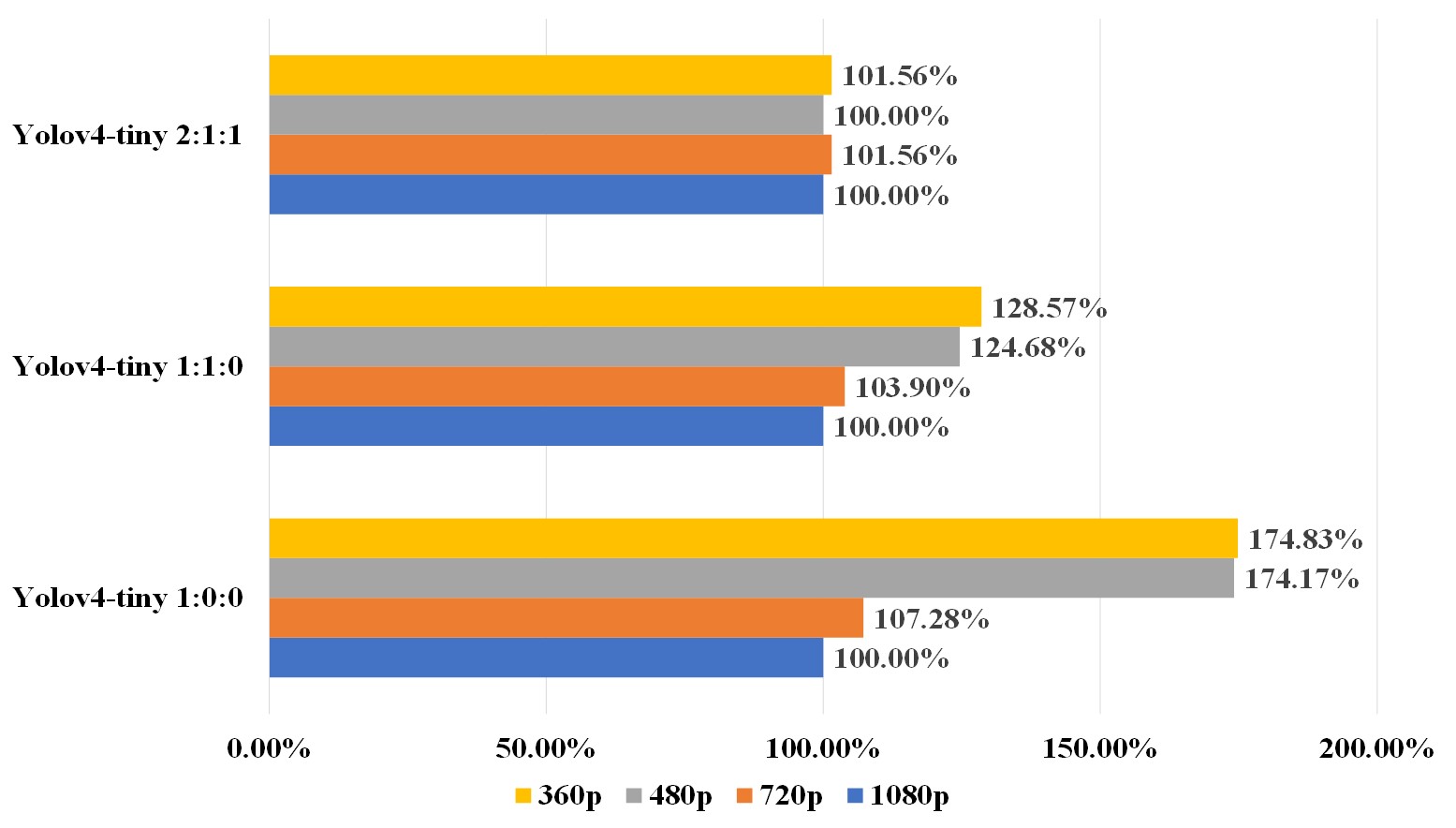}
	\caption{Time growth rate of three mosaic methods that cannot be separated at different resolutions}
\end{figure*}

As can be seen from Table 5, the recognition speed at all four resolutions is low because the Yolov4 network structure is more complex and has more weight parameters, which leads to its higher arithmetic power required for recognition. The recognition speed of Yolov4-tiny algorithm at 360p video resolution is 179.50\% of its recognition speed at 1080p video resolution, and 462.60\% of the recognition speed of Yolov4 algorithm at 360p video resolution.

Since the recognition speed of different algorithms varies at different resolutions, the importance of improving the recognition ability of mosaic at different resolutions is self-evident. Thus, the recognition ability of the overlap problem is judged by comparing the number of frames in the interval from the beginning of the overlap to the end of the overlap in which the Yolov4-tiny algorithm with different mosaic methods correctly identifies and separates the two ships. The specific experimental results are shown in Table 6.

From Table 6, it can be seen that with the decrease of resolution, each algorithm has different degrees of weakening for overlapping target recognition speed, and the degree of weakening can be expressed by the growth rate of the time when it is impossible to separate, as shown in Figure 11. It can be seen that the Yolov4-tiny algorithm using the improved mosaic method achieves a greater advantage at all four different resolutions. Not only the time to fail to separate the overlapping targets is shorter when the ship targets overlap, but also the high recognition rate can be maintained when the resolution degradation occurs.

The experiments prove that the improved mosaic data enhancement method has a certain improvement on the recognition effect of overlapping targets of ships, and the applied Yolov4-tiny algorithm can be deployed on small mobile devices and can be flexibly assembled on various platforms, thus realizing the real-time monitoring of ship targets on the sea surface in the offline state.

\section{Conclusion}

In this paper, an improved mosaic data enhancement method is proposed based on the Yolov4-tiny algorithm as the experimental basis, and the different ratios of mosaic filling are tested for comparison to investigate its effect on the detection accuracy of ship overlapping targets. In the simulation experiments, the ship dataset is analyzed, and the ship overlapping targets in the dataset are used as detection objects. Compared with the original algorithm, the improved method enhances the recognition accuracy of the test dataset by 2.5\% and achieves the same detection effect as the original Yolov4 algorithm, which improves the accuracy and reduces the arithmetic power consumption at the same time. In the real test experiments, the algorithm was deployed on a small mobile testbed for real tests, and the improved method achieved a 17\% reduction in target loss time and a 27.01\% improvement in recognition stability at different video resolutions compared to the original algorithm. Therefore, the target recognition algorithm can improve its ability to recognize overlapping targets of ships after being trained by the improved mosaic data enhancement method.

\section*{Acknowledgment}

This work was supported in part by the National Natural Science Foundation of China under Grant No.51879118, in part by the Natural Science Foundation of Fujian Province No.2020J01688, in part by the Transportation Industry High-Level Technical Talent Training Project No.2019-014, in part by t the Science and Technology Support Project of Fujian Province No. B19101, in part by the Young Talent of Jimei University No. ZR2019006.

\ifCLASSOPTIONcaptionsoff
  \newpage
\fi




\bibliographystyle{IEEEtran}
\bibliography{reference}

\begin{thebibliography}{10}
\providecommand{\url}[1]{#1}
\csname url@samestyle\endcsname
\providecommand{\newblock}{\relax}
\providecommand{\bibinfo}[2]{#2}
\providecommand{\BIBentrySTDinterwordspacing}{\spaceskip=0pt\relax}
\providecommand{\BIBentryALTinterwordstretchfactor}{4}
\providecommand{\BIBentryALTinterwordspacing}{\spaceskip=\fontdimen2\font plus
\BIBentryALTinterwordstretchfactor\fontdimen3\font minus
  \fontdimen4\font\relax}
\providecommand{\BIBforeignlanguage}[2]{{%
\expandafter\ifx\csname l@#1\endcsname\relax
\typeout{** WARNING: IEEEtran.bst: No hyphenation pattern has been}%
\typeout{** loaded for the language `#1'. Using the pattern for}%
\typeout{** the default language instead.}%
\else
\language=\csname l@#1\endcsname
\fi
#2}}
\providecommand{\BIBdecl}{\relax}
\BIBdecl

\bibitem{1}
H.~Guo, X.~Yang, N.~Wang, and X.~Gao, ``A centernet++ model for ship detection
  in sar images,'' \emph{Pattern Recognition}, vol. 112, p. 107787, 2021.

\bibitem{2}
L.~Li, Z.~Zhou, B.~Wang, L.~Miao, and H.~Zong, ``A novel cnn-based method for
  accurate ship detection in hr optical remote sensing images via rotated
  bounding box,'' \emph{IEEE Transactions on Geoscience and Remote Sensing},
  vol.~59, no.~1, pp. 686--699, 2020.

\bibitem{3}
J.~Fu, X.~Sun, Z.~Wang, and K.~Fu, ``An anchor-free method based on feature
  balancing and refinement network for multiscale ship detection in sar
  images,'' \emph{IEEE Transactions on Geoscience and Remote Sensing}, 2020.

\bibitem{4}
J.~Wan, J.~Li, Z.~Lai, B.~Du, and L.~Zhang, ``Robust face alignment by cascaded
  regression and de-occlusion,'' \emph{Neural Networks}, vol. 123, pp.
  261--272, 2020.

\bibitem{5}
P.~N. Chowdhury, P.~Shivakumara, S.~Kanchan, R.~Raghavendra, U.~Pal, T.~Lu, and
  D.~Lopresti, ``Graph attention network for detecting license plates in
  crowded street scenes,'' \emph{Pattern Recognition Letters}, vol. 140, pp.
  18--25, 2020.

\bibitem{6}
T.~Liu, W.~Luo, L.~Ma, J.-J. Huang, T.~Stathaki, and T.~Dai, ``Coupled network
  for robust pedestrian detection with gated multi-layer feature extraction and
  deformable occlusion handling,'' \emph{IEEE transactions on image
  processing}, vol.~30, pp. 754--766, 2020.

\bibitem{7}
R.~Girshick, J.~Donahue, T.~Darrell, and J.~Malik, ``Rich feature hierarchies
  for accurate object detection and semantic segmentation,'' in
  \emph{Proceedings of the IEEE conference on computer vision and pattern
  recognition}, 2014, pp. 580--587.

\bibitem{8}
R.~Girshick, ``Fast r-cnn,'' in \emph{Proceedings of the IEEE international
  conference on computer vision}, 2015, pp. 1440--1448.

\bibitem{9}
S.~Ren, K.~He, R.~Girshick, and J.~Sun, ``Faster r-cnn: towards real-time
  object detection with region proposal networks,'' \emph{IEEE transactions on
  pattern analysis and machine intelligence}, vol.~39, no.~6, pp. 1137--1149,
  2016.

\bibitem{10}
K.~He, G.~Gkioxari, P.~Doll{\'a}r, and R.~Girshick, ``Mask r-cnn,'' in
  \emph{Proceedings of the IEEE international conference on computer vision},
  2017, pp. 2961--2969.

\bibitem{11}
W.~Liu, D.~Anguelov, D.~Erhan, C.~Szegedy, S.~Reed, C.-Y. Fu, and A.~C. Berg,
  ``Ssd: Single shot multibox detector,'' in \emph{European conference on
  computer vision}.\hskip 1em plus 0.5em minus 0.4em\relax Springer, 2016, pp.
  21--37.

\bibitem{12}
J.~Redmon, S.~Divvala, R.~Girshick, and A.~Farhadi, ``You only look once:
  Unified, real-time object detection,'' in \emph{Proceedings of the IEEE
  conference on computer vision and pattern recognition}, 2016, pp. 779--788.

\bibitem{13}
J.~Redmon and A.~Farhadi, ``Yolo9000: better, faster, stronger,'' in
  \emph{Proceedings of the IEEE conference on computer vision and pattern
  recognition}, 2017, pp. 7263--7271.

\bibitem{14}
A.~Farhadi and J.~Redmon, ``Yolov3: An incremental improvement,''
  \emph{Computer Vision and Pattern Recognition, cite as}, 2018.

\bibitem{15}
A.~Bochkovskiy, C.-Y. Wang, and H.-Y.~M. Liao, ``Yolov4: Optimal speed and
  accuracy of object detection,'' \emph{arXiv preprint arXiv:2004.10934}, 2020.

\bibitem{16}
M.~Jiang, Y.~Rao, J.~Zhang, and Y.~Shen, ``Automatic behavior recognition of
  group-housed goats using deep learning,'' \emph{Computers and Electronics in
  Agriculture}, vol. 177, p. 105706, 2020.

\bibitem{17}
Z.~Yu, Y.~Shen, and C.~Shen, ``A real-time detection approach for bridge cracks
  based on yolov4-fpm,'' \emph{Automation in Construction}, vol. 122, p.
  103514, 2021.

\bibitem{18}
S.~Albahli, N.~Nida, A.~Irtaza, M.~H. Yousaf, and M.~T. Mahmood, ``Melanoma
  lesion detection and segmentation using yolov4-darknet and active contour,''
  \emph{IEEE Access}, vol.~8, pp. 198\,403--198\,414, 2020.

\bibitem{19}
S.~Ioffe and C.~Szegedy, ``Batch normalization: Accelerating deep network
  training by reducing internal covariate shift,'' in \emph{International
  conference on machine learning}.\hskip 1em plus 0.5em minus 0.4em\relax PMLR,
  2015, pp. 448--456.

\bibitem{20}
B.~Xu, N.~Wang, T.~Chen, and M.~Li, ``Empirical evaluation of rectified
  activations in convolutional network,'' \emph{arXiv preprint
  arXiv:1505.00853}, 2015.

\bibitem{21}
C.-Y. Wang, H.-Y.~M. Liao, Y.-H. Wu, P.-Y. Chen, J.-W. Hsieh, and I.-H. Yeh,
  ``Cspnet: A new backbone that can enhance learning capability of cnn,'' in
  \emph{Proceedings of the IEEE/CVF conference on computer vision and pattern
  recognition workshops}, 2020, pp. 390--391.

\bibitem{22}
T.-Y. Lin, P.~Doll{\'a}r, R.~Girshick, K.~He, B.~Hariharan, and S.~Belongie,
  ``Feature pyramid networks for object detection,'' in \emph{Proceedings of
  the IEEE conference on computer vision and pattern recognition}, 2017, pp.
  2117--2125.

\bibitem{23}
S.~Yun, D.~Han, S.~J. Oh, S.~Chun, J.~Choe, and Y.~Yoo, ``Cutmix:
  Regularization strategy to train strong classifiers with localizable
  features,'' in \emph{Proceedings of the IEEE/CVF International Conference on
  Computer Vision}, 2019, pp. 6023--6032.

\bibitem{24}
Z.~Zheng, P.~Wang, W.~Liu, J.~Li, R.~Ye, and D.~Ren, ``Distance-iou loss:
  Faster and better learning for bounding box regression,'' in
  \emph{Proceedings of the AAAI Conference on Artificial Intelligence},
  vol.~34, no.~07, 2020, pp. 12\,993--13\,000.

\bibitem{25}
M.~Everingham, L.~Van~Gool, C.~K. Williams, J.~Winn, and A.~Zisserman, ``The
  pascal visual object classes (voc) challenge,'' \emph{International journal
  of computer vision}, vol.~88, no.~2, pp. 303--338, 2010.

\bibitem{26}
T.-Y. Lin, M.~Maire, S.~Belongie, J.~Hays, P.~Perona, D.~Ramanan,
  P.~Doll{\'a}r, and C.~L. Zitnick, ``Microsoft coco: Common objects in
  context,'' in \emph{European conference on computer vision}.\hskip 1em plus
  0.5em minus 0.4em\relax Springer, 2014, pp. 740--755.

\bibitem{27}
Z.~Shao, W.~Wu, Z.~Wang, W.~Du, and C.~Li, ``Seaships: A large-scale precisely
  annotated dataset for ship detection,'' \emph{IEEE transactions on
  multimedia}, vol.~20, no.~10, pp. 2593--2604, 2018.

\end{thebibliography}



\end{document}